\documentclass[letterpaper, 10 pt, conference]{ieeeconf}
\usepackage{times}
\usepackage[pdftex]{graphicx}
\usepackage{caption}
\usepackage{subfig}
\usepackage{tikz}
\usepackage{gensymb}
\usepackage{amsmath}

\usepackage{amsmath,amssymb,amsopn,amstext,amsfonts}
\usepackage{cancel}
\usepackage[space]{cite}
\usepackage{color}
\usepackage{mathtools}
\usepackage{algorithm}
\usepackage{algorithmicx}
\usepackage{algpseudocode}

\usepackage{bm}

\usepackage{diagbox}
\usepackage{booktabs}
\usepackage[linkcolor=black,citecolor=black,urlcolor=black,colorlinks=true]{hyperref}
\usepackage{paralist}
\makeatletter
\def\tagform@#1{\maketag@@@{\normalsize(#1)\@@italiccorr}}
\makeatother

\newcommand{\figref}[1]{Fig.~\ref{#1}}
\newcommand{\tabref}[1]{Table \ref{#1}}
\newcommand{\secref}[1]{Section \ref{#1}}

\graphicspath{{./Figures/}}
\DeclareGraphicsExtensions{.pdf,.png,.jpg,.eps,.svg}
\IEEEoverridecommandlockouts
\title{\LARGE \bf
Head Stabilization for Wheeled Bipedal Robots via Force-Estimation-Based Admittance Control}

\author{Tianyu Wang*, Chunxiang Yan*,  Xuanhong Liao, Tao Zhang, Ping Wang, Cong Wen, \\
Dingchuan Liu, Haowen Yu, Ximin Lyu%
\thanks{This work was supported by the Guangdong-Hong Kong-Macao Joint Research of Science and Technology Planning Funding (Grant No. 2023A0505010019), the National Natural Science Foundation of China (Grant No. 62303495) and the Young Talent Support Project of Guangzhou Association for Science and Technology (Grant No. QT-2025-004).\textit{ (Corresponding author: Ximin Lyu) }}%
\thanks{Tianyu Wang, Cong Wen, Haowen Yu, Ping Wang, Dingchuan Liu and Ximin Lyu are with the School of Intelligent Systems Engineering, Sun Yat-sen University, Guangzhou, 510275, China.}%
\thanks{Chunxiang Yan is with GAC R\&D Center, Guangzhou, 511434, China}
\thanks{Xuanhong Liao is with Direct Drive Technology Limited, Dongguan, 523808, China.}%
\thanks{Tao Zhang is with the Suzhou Nuclear Power Research Institute Co., Ltd, Shenzhen, 518038, China.}
\thanks{*These authors contributed to the work equally and should be regarded as co-first authors}
\thanks{Email:{\tt\small lvxm6@mail.sysu.edu.cn}}
\thanks{For additional demos, watch the included video: https://youtu.be/QzV-891VhtM.}
}

\begin{document}
\maketitle
\begin{abstract}
Wheeled bipedal robots are emerging as flexible platforms for field exploration. However, head instability induced by uneven terrain can degrade the accuracy of onboard sensors (e.g., cameras) or damage fragile payloads. Existing research primarily focuses on stabilizing the mobile platform but overlooks active stabilization of the head in the world frame, resulting in vertical oscillations that undermine overall stability. To address this challenge, we developed a model-based ground force estimation method for our 6-degree-of-freedom (6-DOF) wheeled bipedal robot. Leveraging these force estimates, we implemented an admittance control algorithm to enhance terrain adaptability. Simulation experiments validated the real-time performance of the force estimator and the robot’s robustness when traversing uneven terrain.
\end{abstract}

\section{INTRODUCTION}


As robotics technology advances, wheeled bipedal robots are being increasingly deployed for agile exploration\cite{Marco2022}. However, existing research prioritizes whole-body stabilization while overlooking active regulation of head pose in the world frame. Consequently, terrain-induced oscillations propagate directly to the sensor mast and payload, degrading the accuracy of exteroceptive sensors and jeopardizing the integrity of delicate cargo. Our approach integrates a proprioceptive-only ground contact force estimator with an admittance controller to maintain head-height stability on unstructured terrain.

In recent years, a variety of wheeled bipedal robots (WBRs) with distinct configurations have been developed. In 2017, Boston Dynamics launched the Handle robot, which demonstrated extremely powerful functionality \cite{BD2017handle}, though the technical details of its implementation remain undisclosed. The Ascento robot developed by ETH Zurich \cite{klemm2019ascento, Klemm2020} is driven by four motors, boasting advantages such as a compact structure and low manufacturing costs. Its mobility and exploration capabilities are further constrained by its structural configuration. Meanwhile, it exhibits a deficiency in considering head stability.
SR600 \cite{Zhangchao2019, LiuTangyou2019}, a
hydraulically driven robot designed by the Harbin Institute of
Technology (HIT), utilizes a serial leg configuration with 2
DoFs.
However, its mobility and exploration capabilities are limited, and it fails to maintain head stability when navigating unstructured terrain. The wheeled bipedal robot Ollie, designed by Tencent RoboticX Lab \cite{wang2021balance,Cui2021,zhang2022,zhang2023}, adopts a planar parallel mechanism in its two legs. This structure enhances rigidity and stability but results in a relatively large overall volume of the robot, which, to a certain extent, hinders its exploration ability in confined spaces. 
To further enhance terrain adaptability, Hotwheel \cite{Yu2023} and Nezha \cite{Chen2021} integrate an articulated ankle joint into each leg, resulting in 3-DoF limbs. While this upgrade expands the reachable workspace, it simultaneously increases the number of actuators, exacerbates motor loading.
In our prior work, \cite{liu2024diablo6dofwheeledbipedal}, we designed the 6-DoF Diablo WBR with serial kinematic leg mechanisms, achieving an optimal balance between mobility performance and cost-effectiveness. However, this approach still fails to account for the stability of the head.
Notably, while these robots all emphasize overall dynamics in their design and control, they overlook the critical need for head height stability in the world frame.

At the control technology level, existing wheeled bipedal robots lack active stabilization along the z-axis, relying instead on fixed leg lengths and roll angle control to maintain balance. The conventional PID-controlled leg length locking mechanism fails to preserve head height stability in the world coordinate frame when traversing slopes.

To address these limitations, we propose a novel admittance control framework based on the Wheeled Spring-Damper Inverted Pendulum (W-SDIP) model, enabling simultaneous head stabilization and terrain adaptability.
Traditional admittance control methods, widely used in physical interaction tasks, typically rely on direct force measurements from external sensors such as foot-mounted force/torque sensors\cite{Hogan1984}.
A representative example is Jo and Oh’s work \cite{jo2020}, which achieved stable humanoid locomotion on uneven terrain via impedance control. However, their implementation required explicit force/torque sensor data for contact force measurement.
Similarly, Landi et al. \cite{landi2017} developed an adaptive admittance control framework tailored to human-robot interaction scenarios, but their approach depended on wrist-mounted force/torque sensors.
For hybrid wheeled-legged systems, Tsai et al. \cite{tsai2022} demonstrated successful impedance control deployment, yet their method still relied fundamentally on physical force feedback mechanisms.
Additionally, Chen et al. \cite{Chen2024} proposed a shear-thickening fluid-inspired admittance controller that inherently distinguishes traction from impact without additional force classification, exhibiting enhanced robustness in human-robot collaboration.

Using a simplified dynamic model, this study develops a real-time ground-contact force estimator that, when integrated with admittance control, adjusts leg length to enable terrain adaptation.
This not only enhances adaptability to complex terrain but also ensures head height stability, providing a robust foundation for mounting vision and other precision sensors.
While commercially available six-axis wheel force transducers can directly measure ground-wheel contact forces, they necessitate specialized hardware, introducing additional mass, cost, and installation complexity.
Thus, instead of relying on such extra sensors, this work centers on estimating contact forces using only onboard measurements (without the need for additional sensing hardware).
Notably, the admittance controller functions without external sensors (e.g., LiDAR or force plates).
By estimating external forces exclusively from the robot’s dynamic model, the method reduces hardware costs, eliminates traditional reliance on sophisticated sensors, and paves a low-cost, high-performance path toward widespread adoption of wheeled bipedal robots.

The main contributions of this paper are as follows:
\begin{itemize}
\item We develop two robot dynamics models: a 2-DoF planar leg model for real-time wheel-ground contact force calculation, and a W-SDIP whole-body model to inform admittance control design. (\secref{sec:modeling})
\item We develop a force estimator based on the dynamics of a 2-DoF planar Leg model. (\secref{sec:estimation})
\item We design an adaptive admittance control framework that leverages estimated contact forces to regulate leg motion. (\secref{sec:control})
\end{itemize}

\section{MODELING}\label{sec:modeling}

In this work, we develop two models to enable head-height stabilization. First, we model each leg as a 2-DoF planar serial linkage for real-time ground contact force estimation. Second, we formulate the entire robot as a wheeled spring-damper inverted pendulum (W-SDIP), treating the legs as compliant elements to adapt terrain disturbances via admittance shaping rather than high-gain feedback. The 2-DoF Planar Leg Model serves as the physical foundation for the force estimator, while the W-SDIP model forms the basis for the admittance controller.
Such a decoupled approach allows for more targeted modeling of the robot’s different functional requirements, thereby enabling the corresponding modules to better achieve their respective objectives.

\subsection{2-DoF Planar Leg Model}

\begin{figure}[t]
    \centering
     \includegraphics[width=0.9\columnwidth]{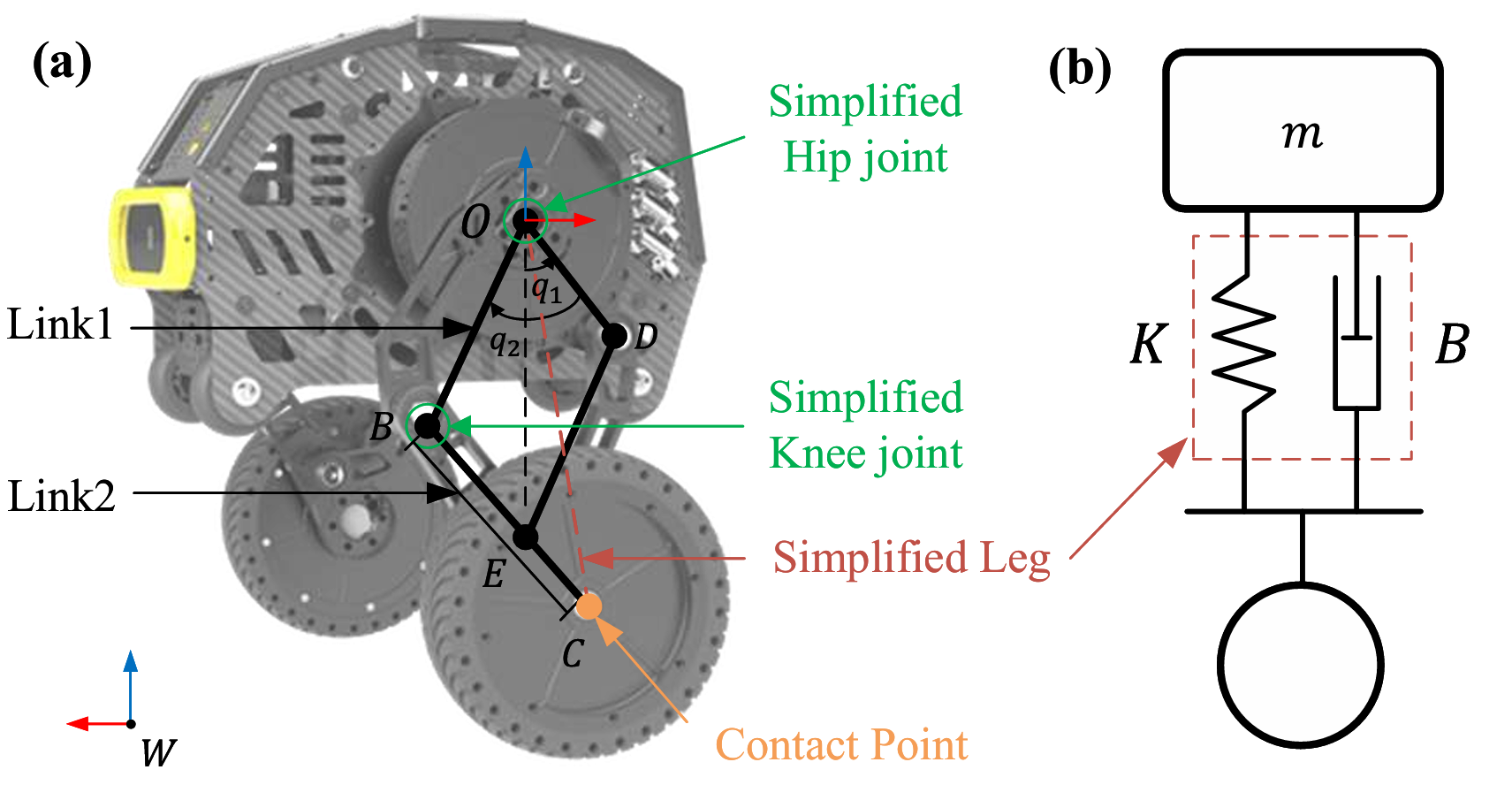}
    \caption{\textbf{Robot coordinate systems, generalized coordinates, actuated torques, and leg structure.} } 
    \label{fig:leg_mechanism}
    \vspace{-0.5cm}
\end{figure}

\figref{fig:leg_mechanism}(a) depicts the simplified leg used for contact force estimation.  
As illustrated in \figref{fig:leg_mechanism}(a), we define the base frame \(\{O\}\) and the world frame \(\{W\}\). In this paper, the red and blue arrows represent the $x$ and $z$ axes of a coordinate system, respectively.
Point $O$ is the joint connecting links $OB$ and $OD$, and serves as the fixed base of the 2-DoF planar leg model.
We use this model to separately model the two legs, applying the same approach to both.

The leg mainly comprises two serial links: 

\begin{itemize}
  \item Link1($OB$), length $L$, and is parallel to link $DE$.
  \item Link2($BC$), length $L$, and is parallel to link $OD$.
\end{itemize}

As illustrated in \figref{fig:leg_mechanism}(a), we define the simplified hip and knee joint structures in the simplified model, which will be referred to as the hip joint (located in point $O$) and the knee joint (located in point $B$) hereinafter. 
The configuration is fully described by two angles measured counter-clockwise:
\begin{itemize}
  \item $q_{1}$ -- angle from the negative $z$-axis to link $OD$.
  \item $q_{2}$ -- angle from link $OD$ to link $OB$.
\end{itemize}
In this paper, we define clockwise angles as negative and counterclockwise angles as positive. Hence, the angle from the negative $z$-axis to link $OB$ is $q_{1}+q_{2}$. 
The Cartesian coordinates of the wheel center are
\begin{align}
x &= L\sin q_{1} + L\sin(q_{1}+q_{2}),\\
z &= -L\cos q_{1}  -L\cos(q_{1}+q_{2}).
\end{align}
The geometric Jacobian relating joint velocities $\dot{\mathbf{q}}=[\dot{q}_{1},\dot{q}_{2}]^T$ to the linear velocity $\mathbf{v}_{W}=[\dot{x},\dot{z}]^T$ is
\begin{equation}\label{eq:Jacobian}
\boldsymbol{J}(\boldsymbol{q}) =
\begin{bmatrix}
L(\cos q_{1}+\cos(q_{1}+q_{2})) & L\cos(q_{1}+q_{2})\\[4pt]
\,L(\sin q_{1}+\sin(q_{1}+q_{2})) & \,L\sin(q_{1}+q_{2})
\end{bmatrix}.
\end{equation}


\begin{figure*}[th]
	\centering
        \includegraphics[width=0.90\textwidth]{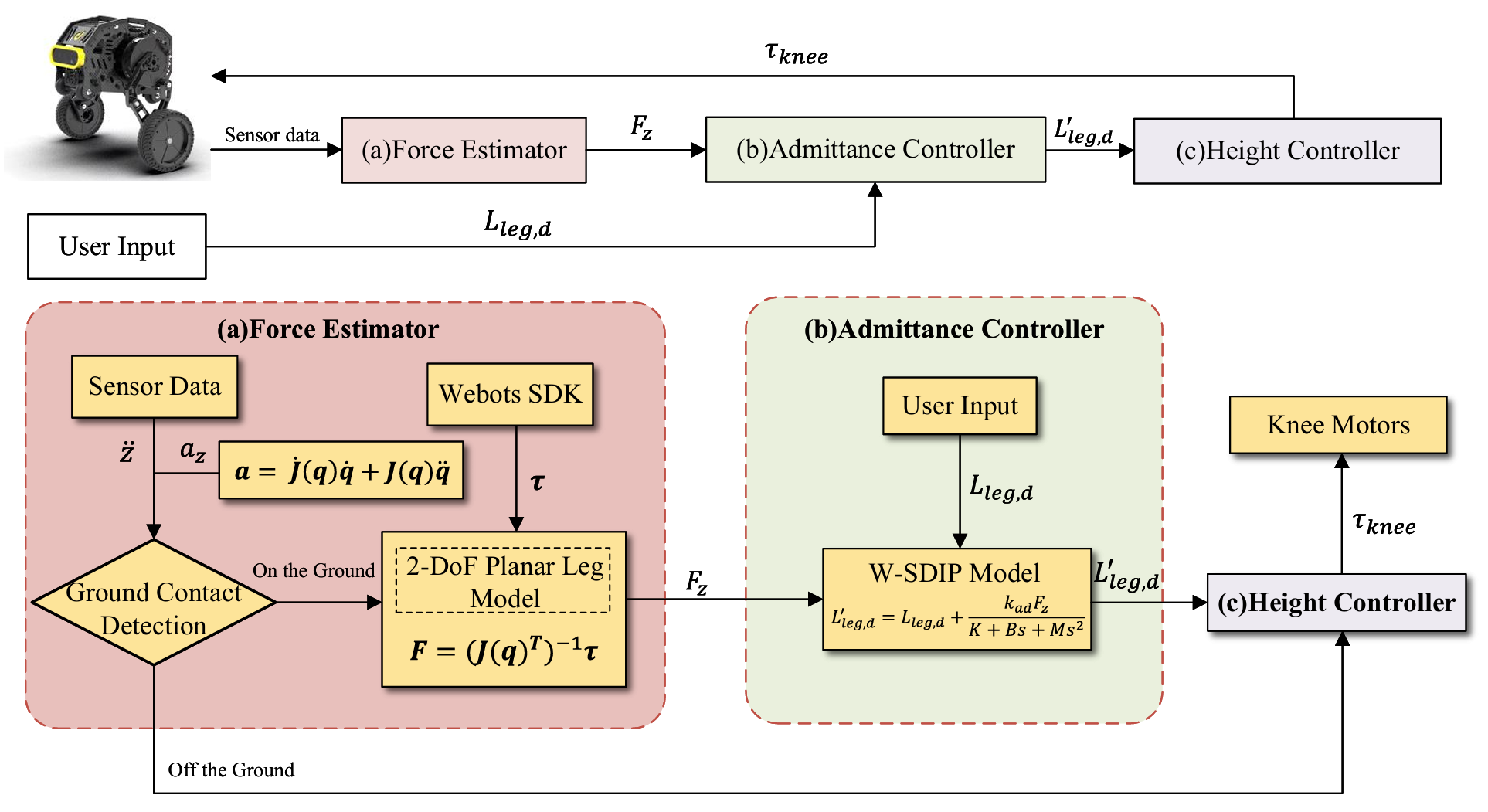}
        \caption{\textbf{Admittance control Framework}  }
        \label{fig:Admittance_framework}
        \vspace{-0.5cm}
\end{figure*}
\subsection{Wheeled Spring–Damper Inverted Pendulum (W-SDIP) Model}

\label{subsec:wsdip}


As shown in \figref{fig:leg_mechanism} (b), the robot is modeled as a point mass $m$ at the hip, connected to the ground contact point via a massless leg. This leg acts as a spring-damper pair: its restoring force is proportional to the deviation between the instantaneous leg length and the nominal length (with proportionality governed by the stiffness coefficient $K$) and to the rate of this length change (with proportionality governed by the damping coefficient $B$).

The W-SDIP model’s design allows the robot to adapt to diverse environmental and task demands via parameter tuning of the spring and damper. This adaptability is critical for realizing flexible, stable motion control.

\section{ADMITTANCE CONTROL WITH GROUND CONTACT FORCE ESTIMATION}\label{sec:control}

As shown in \figref{fig:Admittance_framework}, our admittance control framework integrates two functionally coupled modules:
(a) Force estimator module:
This module continuously fuses sensor streams (including joint encoders and an IMU), models each leg of the wheeled bipedal robot as a 2-DoF manipulator, and determines whether the robot is off the ground and the contact force magnitude.
(b) Admittance controller module:
The operator specifies a nominal target height~$L_{\text{leg,d}}$.
Leveraging the estimated contact force~$F_z$ from the force estimator module,
the admittance controller dynamically adjusts this reference to yield an
updated target height~$L'_{\text{leg,d}}$.
Subsequently, the height controller computes the knee torque command $\tau_{\text{knee}}$ to track $L'_{\text{leg,d}}$.

\subsection{Force Estimator Module}\label{sec:estimation}

The goal of the force estimator module is to quantify the normal support force exerted by the terrain on each wheel—specifically, the ground reaction force perpendicular to the contact surface. 

The force estimation procedure is summarized below.
\begin{enumerate}
  \item Perform a ground contact detection; 
  \item If the wheel is in contact with the ground, compute the contact force $F_z$;
  \item If the wheel is off the ground, run the height controller directly.
  
\end{enumerate}

\subsubsection{Ground contact detection}
A preliminary check is carried out by comparing the vertical acceleration measured at the robot’s head (in the world frame \(\{W\}\)) with the acceleration of the wheels (in the 2-DoF planar leg model's base frame \(\{O\}\)).
Notably, our accelerometer is mounted directly on the robot's head, such that the measured acceleration \(\ddot{\boldsymbol{z}}\) corresponds to the acceleration of the head in the world coordinate system.
Additionally, we can calculate the acceleration between the wheel center of the robot and the head using the robot’s joint angles, angular velocities, and angular accelerations. This is enabled by the acceleration mapping formula from joint space to operational space:
\begin{align}
\boldsymbol{a} = \dot{\boldsymbol{J}}(\boldsymbol{q})\boldsymbol{\dot{q}} + \boldsymbol{J}(\boldsymbol{q})\boldsymbol{\ddot{q}},
\end{align}
where \( \boldsymbol{J}(\boldsymbol{q}) \) is the Jacobian matrix relating joint angles \( \boldsymbol{q} \) to the wheel center motion, $\boldsymbol{a}=[a_x,\,a_z]^T$ comprises the components of the wheel center acceleration along the $x$- and $z$-axes of frame \(\{O\}\), respectively.

As a prerequisite for accurate force estimation, we first determine whether the robot maintains ground contact.  
We discretize the two-dimensional acceleration space $(a_z,\ddot{z})$ into nine exhaustive and mutually exclusive cases, summarised in Table~\ref{tab:contact_logic}.  

The robot is highly likely to be experiencing external lifting forces and in an off-ground state when any of the following kinematic signatures are observed:
\begin{itemize}
    \item \textbf{$\ddot{z} > g$}, \textbf{$a_{z} > 0$}: Concurrent leg retraction and upward head acceleration indicates active lifting, typically occurring when the robot is being picked up.
    \item \textbf{$\ddot{z} > g$}, \textbf{$a_{z} = 0$}: Sustained upward head acceleration without leg movement suggests the robot is being picked up while maintaining a static posture.
    \item \textbf{$\ddot{z} = g$}, \textbf{$a_{z} > 0$}: Leg retraction while maintaining near-zero head acceleration implies the robot is passively supported, such as being placed on a platform.
    \item \textbf{$\ddot{z} = g$}, \textbf{$a_{z} < 0$}: Leg extension with negligible head acceleration similarly indicates passive support conditions.
\end{itemize} 

The robot is highly likely to be in a free-fall state (off-ground) when any of the following kinematic signatures are observed:
\begin{itemize}
    \item \textbf{$\ddot{z} < g$}, \textbf{$a_{z} <= 0$}: Concurrent leg extension and downward head acceleration indicate a free-fall state.
\end{itemize} 

In all remaining states where the contact condition cannot be definitively determined, the system defaults to assuming ground contact and proceeds with subsequent force estimation procedures.

Using~$\ddot{z}$ and~$a_z$, we develop a concise set of \emph{loss-of-contact detection} rules. 
Whenever any rule is satisfied, the robot is deemed to have left the ground, at which point the system immediately disables both the force estimator and subsequent admittance control while switching directly to the height controller.

\begin{table}[h]
\centering
\caption{Ground contact detection based on combinations of base acceleration $\ddot{z}$ and wheel-relative acceleration $a_{z}$}
\label{tab:contact_logic}
\begin{tabular}{@{}cccc@{}}
\toprule
\textbf{\ $\ddot{z}$} & \textbf{ $a_{z}$} & \textbf{Contact Status} & \textbf{Force Estimation} \\ 
\midrule
$\ddot{z} > g$ & $a_{z} > 0$ & off the ground & Skip \\
$\ddot{z} > g$ & $a_{z} < 0$ & off the ground & Required \\
$\ddot{z} > g$ & $a_{z} = 0$ & on the ground & Skip \\[4pt]
$\ddot{z} < g$ & $a_{z} > 0$ & on the ground & Required \\
$\ddot{z} < g$ & $a_{z} < 0$ & off the ground & Skip \\
$\ddot{z} < g$ & $a_{z} = 0$ & off the ground & Skip \\[4pt]
$\ddot{z} = g$ & $a_{z} > 0$ & off the ground & Skip \\
$\ddot{z} = g$ & $a_{z} < 0$ & off the ground & Skip \\
$\ddot{z} = g$ & $a_{z} = 0$ & on the ground & Required \\
\bottomrule
\end{tabular}
\end{table}
When continued ground contact is confirmed, the estimator proceeds with the computation of the contact load.

\subsubsection{The Computation of Contact Force}
During motion, the virtual work performed by the joint torques equals that performed by the ground contact force. That is,
\begin{align}\label{tau}
\boldsymbol{\tau}^T \delta \boldsymbol{q} = \boldsymbol{F}^T \delta \boldsymbol{x},
\end{align}
where \(\delta \boldsymbol{q}\) is the virtual displacement vector of the joints (infinitesimal change in joint angles), and \(\delta \boldsymbol{x}\) is the virtual operational displacement vector of the contact point.
The 2-DoF planar leg model employed in this work omits the geometric and inertial properties of the driven wheel present in the physical system. This simplification creates a critical dynamic gap: the external moment arising from ground friction, transmitted through the wheel-ground contact point, must be incorporated into the joint space to maintain model fidelity. To rigorously account for this wheel-induced dynamic coupling, we introduce continuous mapping coefficients $k_1$ and $k_2$. These coefficients project the torque $\tau_{\mathrm{wheel}}$ generated by the wheel motor onto the joint space as equivalent disturbance torques. Both \(k_1\) and \(k_2\) are hyperparameters that require tuning during experiments.Thus, the total joint torque vector $\boldsymbol{\tau}$ is given by:
\begin{align}
\boldsymbol{\tau} &= 
\begin{bmatrix}
\tau_{\mathrm{hip}} + k_{1}\tau_{\mathrm{wheel}} \\[4pt]
\tau_{\mathrm{knee}} + k_{2}\tau_{\mathrm{wheel}}
\end{bmatrix},
\end{align}
where $\tau_{\mathrm{hip}}$ and $\tau_{\mathrm{knee}}$ denote the nominal actuation torques at the hip and knee joints, respectively. This projection mechanism compensates for the omitted wheel dynamics, preserving dynamic equivalence between the simplified model and the full physical system.
By definition of the Jacobian, the virtual end-effector displacement relates to the virtual joint displacement as
\begin{align}\label{virtual}
\delta \boldsymbol{x} = \boldsymbol{J(q)} \cdot \delta \boldsymbol{q},
\end{align}
substitute~\eqref{virtual} into~\eqref{tau},which yields:
\begin{align}
\boldsymbol{F} = (\boldsymbol{J(q)}^T)^{-1} \cdot \boldsymbol{\tau}.
\end{align}

To determine the vertical contact force, the impact of the robot's head motion in the vertical direction on contact force estimation must be taken into account. The contact force of the robot is given by:  
\begin{equation}
F_z = f_z - m\ddot{z},
\end{equation}
where \( m \) represents the mass of the robot head involved in the acceleration motion, 
\( g \) is the gravitational acceleration.

\subsection{Admittance Controller Module}\label{sec: admittance control block}

Admittance control confers spring-damper characteristics to robots during environmental interaction, enabling adaptive motion across uneven terrain. To leverage this capability in our system, we integrate an admittance controller upstream of the height controller. This module processes two inputs: (1) the original desired leg length $L_{\text{leg},d}$ and (2) the estimated external contact force $F_z$ derived from our previously established force estimation framework. The admittance controller generates a modified reference trajectory (expressed as an adjusted leg length $L'_{\text{leg},d}$) that supersedes the original command. This compensated reference is fed to the height controller, which computes the required knee joint torque command. Finally, this torque signal drives the knee joint motor, thereby integrating environmental adaptability into our locomotion control architecture.

The basic principle formula of admittance control is as follows:
\begin{equation}
\begin{aligned}
(L'_{\text{leg},d} - L_{\text{leg},d})K
+ (\dot{L}'_{\text{leg},d} - \dot{L}_{\text{leg},d})B \\
+ (\ddot{L}'_{\text{leg},d} - \ddot{L}_{\text{leg},d})M
= k_{ad}F_z,
\end{aligned}
\label{eq:long}
\end{equation}
where $K$ is the stiffness coefficient, which determines the degree of change of the robot's leg when subjected to external forces. $B$ is the damping coefficient, related to the change speed of the robot's leg, and affects the robot's absorption and damping effect of external forces. $M$ is the inertia coefficient, related to the acceleration of the robot's leg length change, and affects the robot's acceleration response to external forces. The compensation coefficient $k_{ad}$ adjusts the leg-shortening magnitude.


To implement this relationship in the robot’s main control code—where operations execute in discrete time steps—discretization of~\eqref{eq:long} is necessary. This paper employs the backward difference method for discretization.
The first and second derivatives of \(L_{\text{leg},d}\) are approximated as:


\begin{equation}
\dot{L}_{\text{leg},d}[k] \approx \frac{L_{\text{leg},d}[k] - L_{\text{leg},d}[k - 1]}{T},
\end{equation}
\begin{equation}
\ddot{L}_{\text{leg},d}[k] \approx\frac{\dot{L}_{\text{leg},d}[k] - \dot{L}_{\text{leg},d}[k - 1]}{T} ,
\end{equation}
where \( k \) is the current sampling period and \( T \) is the sampling time. 
The adjusted leg length \( L'_{\text{leg},d}[k] \) is computed as:
\begin{equation}
    L'_{\text{leg},d}[k] = L_{\text{leg},d}[k] + \Delta L[k],
\end{equation}
where the length adjustment \(\Delta L[k]\) follows the second-order difference equation:
\begin{equation}
    A_0 \Delta L[k] + A_1 \Delta L[k-1] + A_2 \Delta L[k-2] = k_{ad}T^2 F_z,
\end{equation}
with coefficients:
\begin{align*}
    A_0 &= M + BT + KT^2 \\
    A_1 &= -(2M + BT) \\
    A_2 &= M
\end{align*}

\subsection{Height controller Module}
The height controller's input is the desired leg lengths of the left and right legs, and its output is the torque of the robot's knee joint motors. The principal formula of the robot's height controller is as follows~\cite{liu2024diablo6dofwheeledbipedal}:
\begin{equation}
\Delta F = \ddot{L}_{\text{leg},d} + k_p (L_{\text{leg},d} - L_{\text{leg}}) + k_d (\dot{L}_{\text{leg},d} - \dot{L}_{\text{leg}}),
\end{equation}
\begin{equation}
\tau_{\text{knee}} = J^T (\Delta F + m_H g),
\end{equation}
where \(\Delta F\) is the output force of the combined PD and feedforward controller to adjust height changes in user input. \(\ddot{L}_{\text{leg},d}\), \(\dot{L}_{\text{leg},d}\), and \(L_{\text{leg},d}\) are the desired leg length change acceleration, desired leg length change rate, and desired leg length, respectively.  \(k_d\) and \(k_p\) are the parameters of the PD controller.  \(L_{\text{leg}}\) and \(\dot{L}_{\text{leg}}\) are the estimated leg length and its derivative, respectively. \(\tau_{\text{knee}}\) is the initial torque output acting on the knee joint motor; \(J^T\) is the Jacobian matrix that maps the lifting force of the head to the torque of the knee joint motor.  \(m_H\) is the mass of the robot's head.

\section{EXPERIMENTS}\label{sec:experiments}
All experiments are conducted in Webots~2023a, a high-fidelity open-source robotics simulator that provides accurate rigid-body dynamics, realistic contact forces and a customizable terrain generator.  
Our 6-DoF wheeled bipedal robot DIABLO is modeled at full scale with identical inertial and geometric parameters to the real prototype.  Two challenging terrain scenarios—bilateral slopes and continuous ascending and descending
slopes.
\subsection{Experiment 1: Single Slope Terrain}
The goal of Experiment 1 is to quantify how single slope terrain disturbances affect head stability when the proposed method is enabled versus the baseline, where the baseline is the controller developed in our previous work\cite{liu2024diablo6dofwheeledbipedal}.

\begin{figure}[t]
    \centering
     \includegraphics[width=1.0\columnwidth]{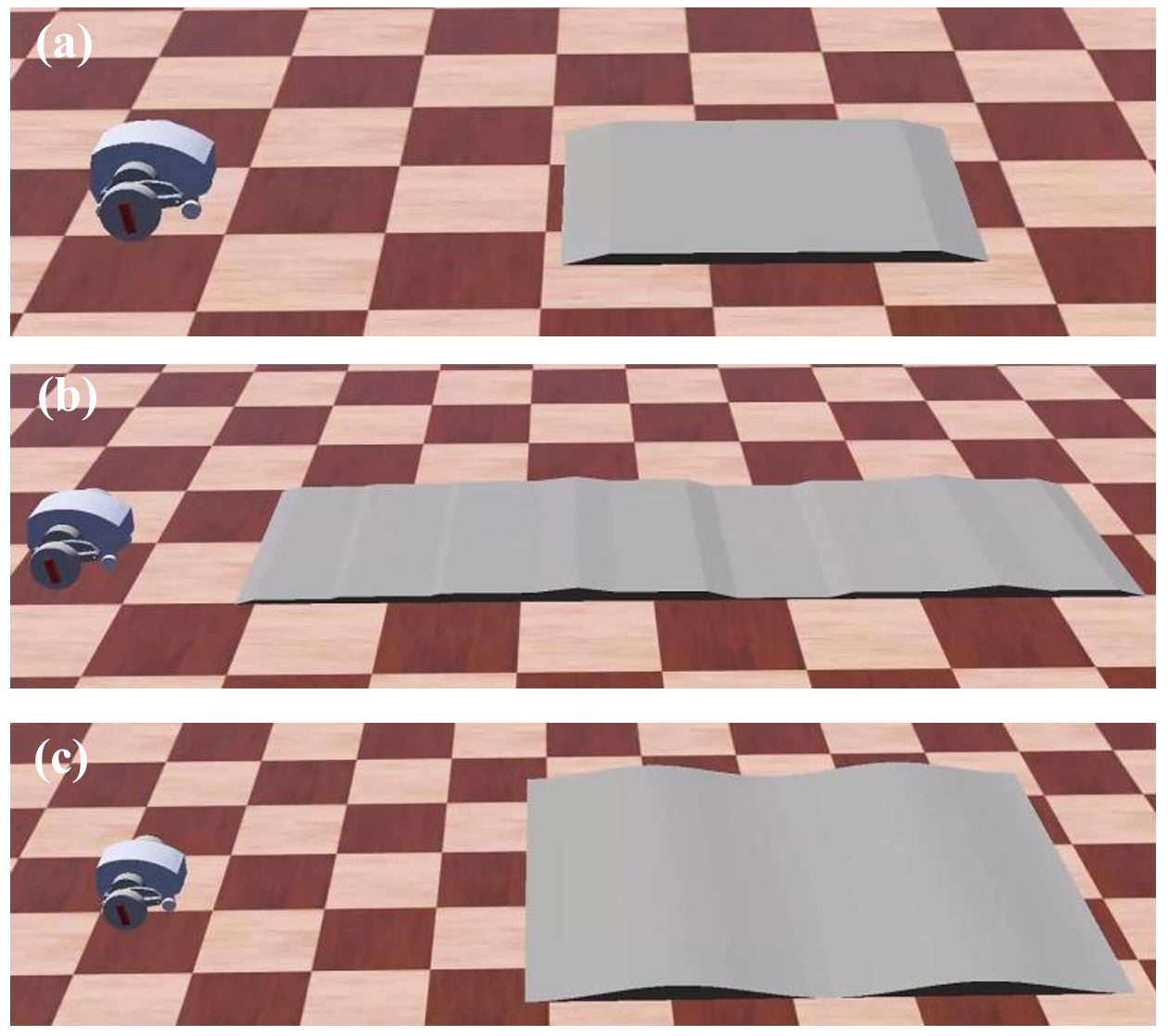}
    \caption{\textbf{Webots simulation environment} (a) Experiment 1: Single Slope Terrain; (b) Experiment 2: High-frequency Rugged Terrain; (c) Experiment 3: Continuous Undulating Terrain;} 
    \label{fig:exp}
    \vspace{-0.3cm}
\end{figure}

As shown in \figref{fig:exp}(a), the terrain consists of a 10° upward slope, followed by a 1 m flat section, and then a 10° downward slope. The robot traverses the terrain at 0.5 m/s while maintaining a constant desired leg length $L_{\text{leg},d}$.

To objectively assess the height-stabilization performance of the proposed method, we adopt three frequently-used metrics: mean absolute error (MAE), root-mean-square error (RMSE) and peak-to-peak amplitude (P2P). 
These metrics respectively quantify the average offset from the reference height, the dispersion of the signal, and the maximum fluctuation range. Improvement percentages are calculated as relative reduction ratios: $\text{improvement} = 100\% \times (1 - \frac{\text{Proposed}}{\text{Baseline}})$.

\begin{figure}[t]
    \centering
     \includegraphics[width=1.0\columnwidth]{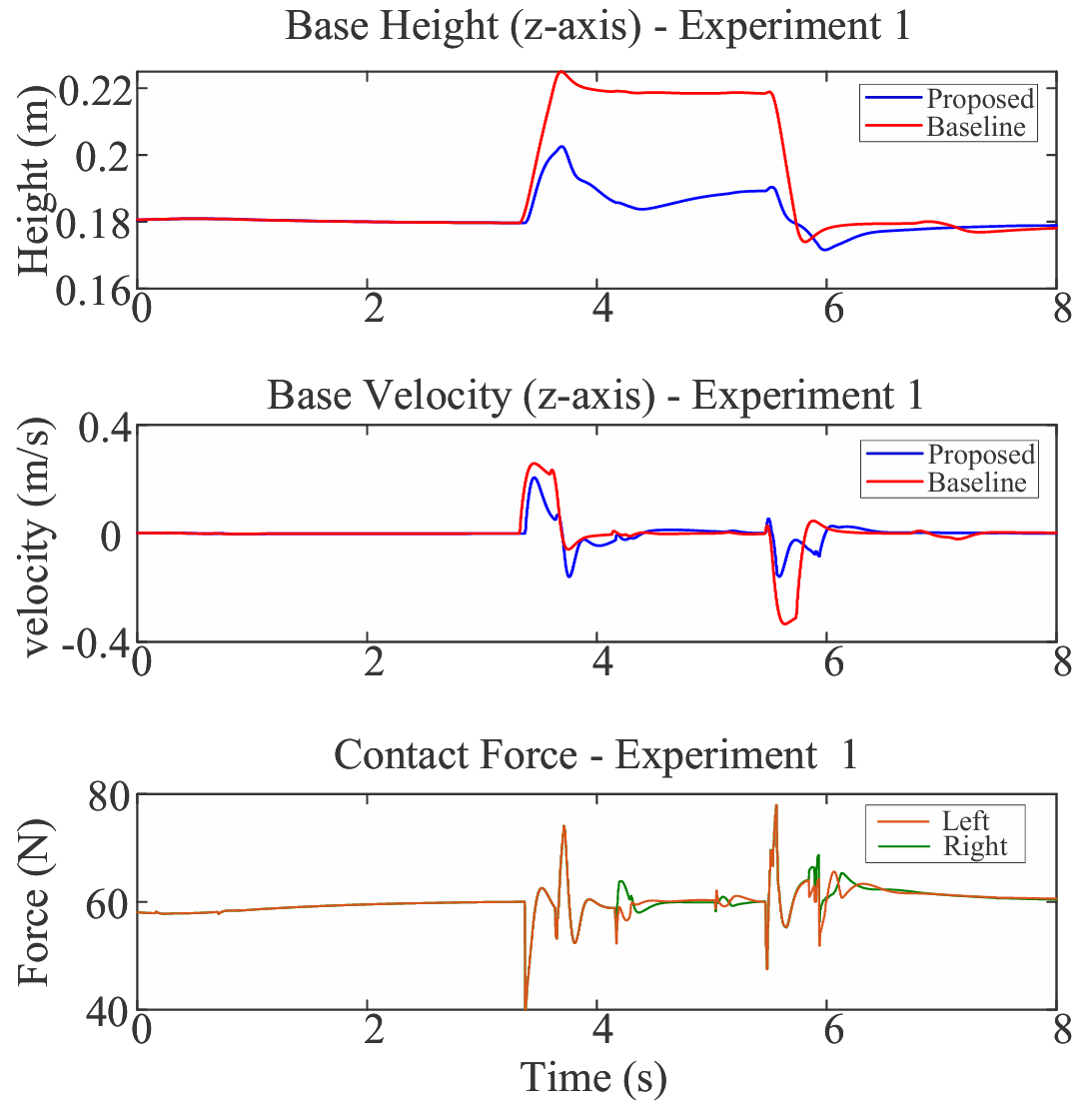}
    \caption{\textbf{Experiment 1: The vertical displacement and velocity of the robot's head along the z-axis, along with concurrent changes in contact forces} } 
    \label{fig:exp_data_1}
    \vspace{-0.1cm}
\end{figure}

\begin{figure}[t]
    \centering
     \includegraphics[width=1.05\columnwidth]{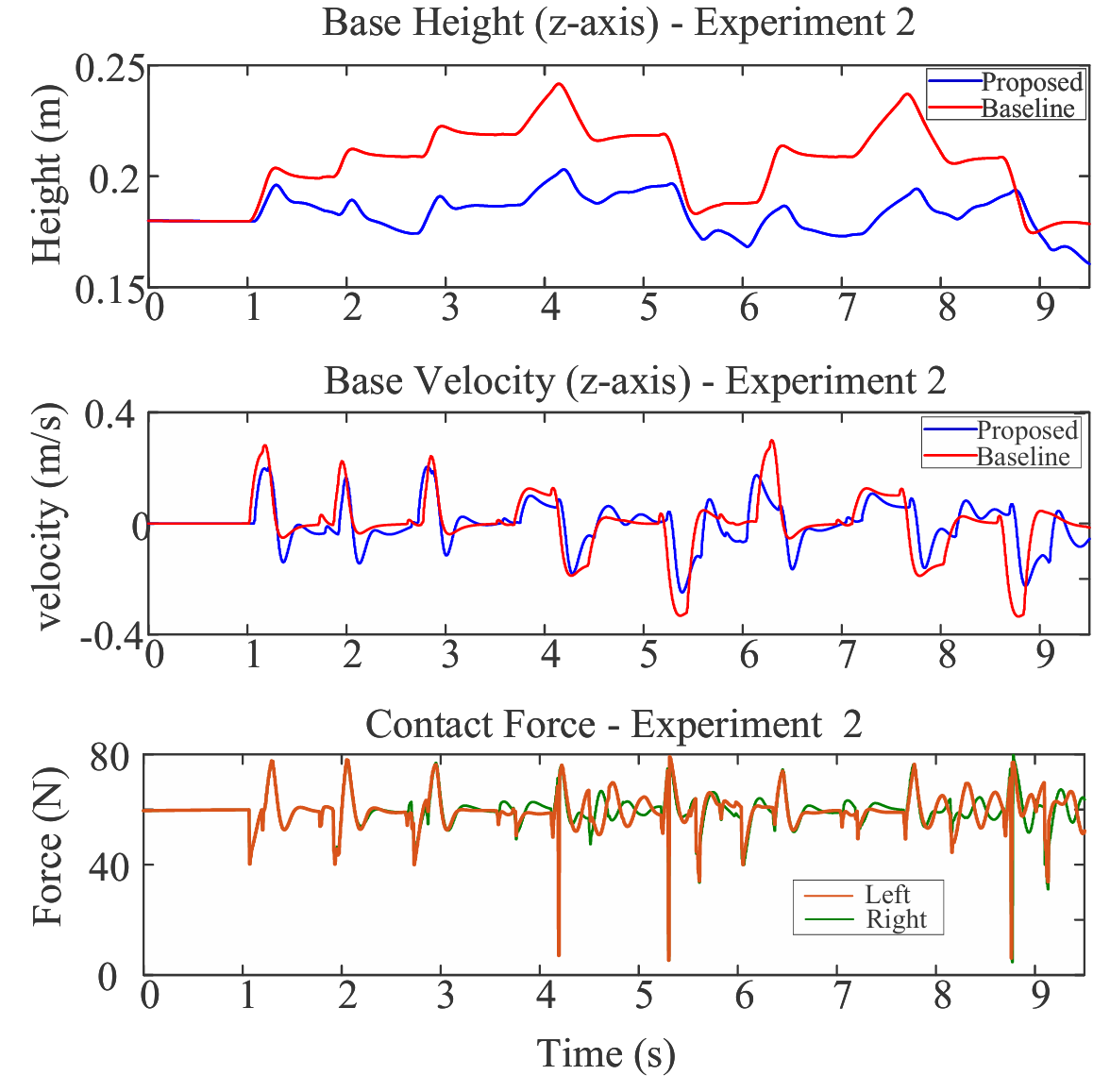}
    \caption{\textbf{Experiment 2: The vertical displacement and velocity of the robot's head along the z-axis, along with concurrent changes in contact forces} } 
    \label{fig:exp_data_2}
    \vspace{-0.1cm}
\end{figure}

As shown in Figure \ref{fig:exp_data_1}, the vertical displacement and velocity of the robot's head along the z-axis, as well as the concurrent changes in contact forces during Experiment 1, are presented. The experimental results are summerized in~\tabref{tab:stability_metrics_1}, the experimental results demonstrate that the proposed method achieves: (1) 68.9\% improved steady-state position accuracy (MAE reduction from 0.0103m to 0.0032m), (2) 65.7\% better dynamic error suppression (RMSE decrease from 0.0143m to 0.0049m), and (3) 35.7\% enhanced extreme disturbance adaptation (peak-to-peak position deviation adaptation from 0.0625m to 0.0402m), while improving velocity stability  (velocity MAE change from 0.0157m/s to 0.0144m/s) with 34.2\% smoother transient responses (velocity RMSE reduction from 0.0538m/s to 0.0354m/s) and 38.2\% lower peak velocity variations (0.5928m/s to 0.56 60m/s).

\begin{table}[h]
\centering
\caption{Stabilization performance metrics on experiment 1.}
\label{tab:stability_metrics_1}
\begin{tabular}{lccccc}
\toprule
\textbf{Metric} & \textbf{Unit} & \textbf{Baseline} & \textbf{Proposed} & \textbf{Improvement} \\
\midrule
\multicolumn{5}{l}{\textit{Position (z)}} \\
MAE & m & 0.0103 & \textbf{0.0032} & \textbf{68.9\%} \\
RMSE & m & 0.0143 & \textbf{0.0049} & \textbf{65.7\%} \\
P2P & m & 0.0625 & \textbf{0.0402} & \textbf{35.7\%} \\

\multicolumn{5}{l}{\textit{Velocity ($v_z$)}} \\ 
MAE & m/s & 0.0157 & \textbf{0.0144} & \textbf{8.3\%} \\
RMSE & m/s & 0.0538 & \textbf{0.0354} & \textbf{34.2\%} \\
P2P & m/s & 0.5928 & \textbf{0.3660} & \textbf{38.3\%} \\
\bottomrule
\end{tabular}
\end{table}

These results demonstrate that the proposed controller significantly improves head-height stability, exhibiting lower bias, reduced variance, and smaller extreme excursions.
\subsection{Experiment 2: High-frequency Rugged Terrain}


The goal of Experiment 2 is to assess how the disturbances caused by high-frequency terrain variations affect head stability by comparing the proposed method's performance with the baseline.

As shown in \figref{fig:exp}(b), this experiment uses a series of high-frequency Rugged terrains to evaluate the robot's head stability maintenance capability when traversing rapidly changing terrains. The robot moves through these terrain profiles at 0.5 m/s while keeping a constant desired leg length $L_{\text{leg},d}$. 
The evaluation analyzes the robot's base height and vertical velocity in world coordinates, using MAE, RMSE, and peak-to-peak values as performance metrics. 





\begin{table}[h]
\centering
\caption{Stabilization performance metrics on experiment 2.}
\label{tab:stability_metrics_2}
\begin{tabular}{lccccc}
\toprule
\textbf{Metric} & \textbf{Unit} & \textbf{Baseline} & \textbf{Proposed} & \textbf{Improvement} \\
\midrule
\multicolumn{5}{l}{\textit{Position (z)}} \\
MAE & m & 0.0169 & \textbf{0.0051} & \textbf{69.8\%} \\
RMSE & m & 0.0185 & \textbf{0.0064} & \textbf{65.4\%} \\
P2P & m & 0.0793 & \textbf{0.0406} & \textbf{48.8\%} \\

\multicolumn{5}{l}{\textit{Velocity ($v_z$)}} \\
MAE & m/s & 0.0422 & \textbf{0.0377} & \textbf{10.7\%} \\
RMSE & m/s & 0.0839 & \textbf{0.0623} & \textbf{25.7\%} \\
P2P & m/s & 0.6357 & \textbf{0.4530} & \textbf{28.7\%} \\
\bottomrule
\end{tabular}
\end{table}
As shown in Figure \ref{fig:exp_data_2}, the vertical displacement and velocity of the robot's head along the z-axis, as well as the concurrent changes in contact forces during Experiment 2, are presented. The experimental results are summerized in~\tabref{tab:stability_metrics_2}, the experimental results demonstrate that the proposed method achieves: (1) 69.8\% improved steady-state position accuracy 
, (2) 65.4\% better dynamic error suppression 
, and (3) 48.8\% enhanced extreme disturbance adaptation
, while improving velocity tracking precision (velocity MAE reduction from 0.0422\,m/s to 0.0377\,m/s, 10.7\%) with 25.7\% smoother transient responses
and 28.7\% lower peak velocity variations
, collectively confirming superior terrain adaptation performance.


\begin{figure}[t]
    \centering
     \includegraphics[width=0.95\columnwidth]{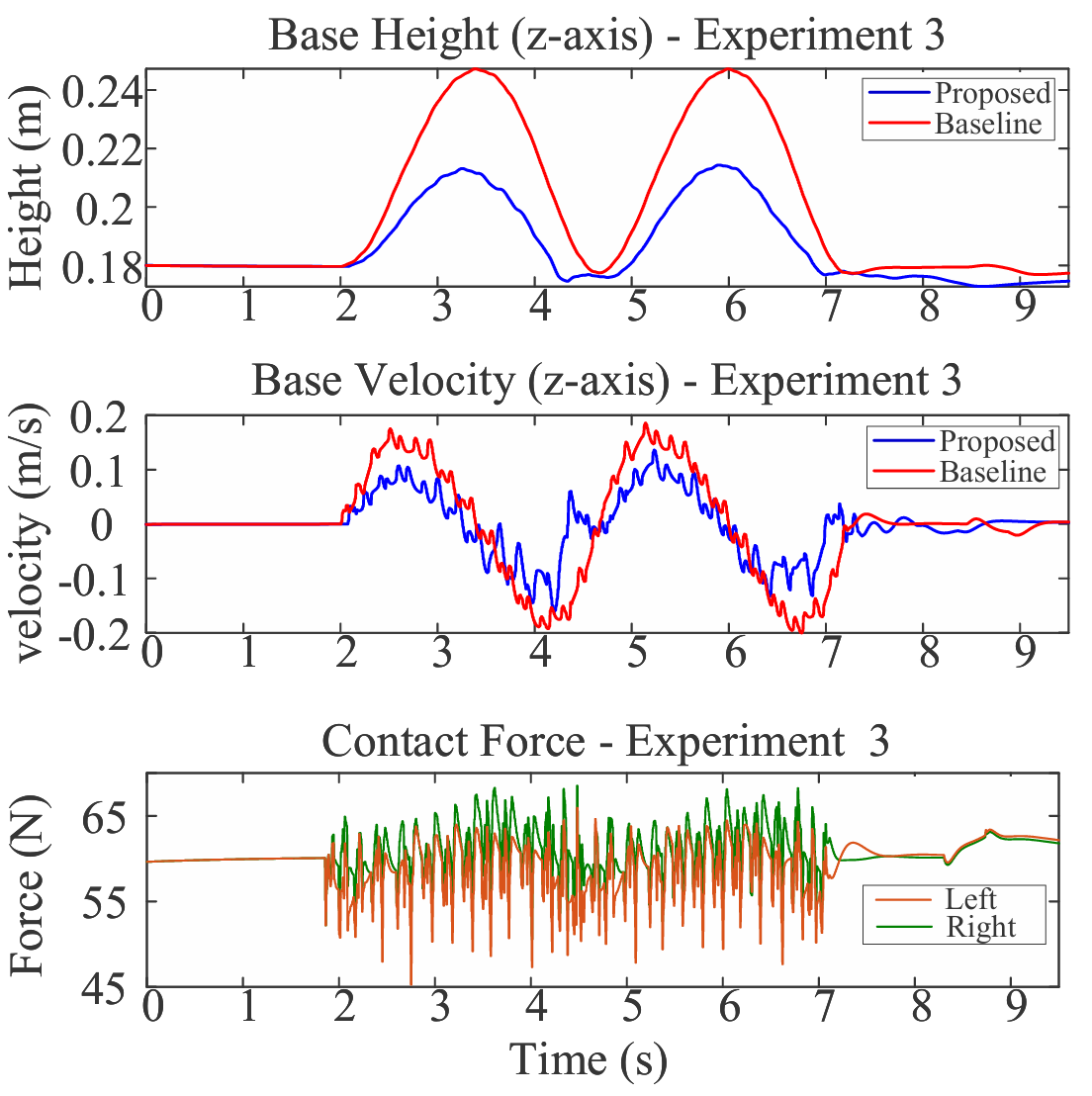}
    \caption{\textbf{Experiment 3: The vertical displacement and velocity of the robot's head along the z-axis, along with concurrent changes in contact forces} } 
    \label{fig:exp_data_3}
    \vspace{-0.1cm}
\end{figure}

\subsection{Experiment 3: Continuous Undulating Terrain}

The goal of Experiment 3 is to verify the robot's adaptive motion performance when traversing continuously undulating terrains, by comparing the proposed method's performance with the baseline.

As shown in \figref{fig:exp}(c), to verify the robot's adaptive motion performance when traversing continuously varying terrains, a sinusoidal terrain was adopted in the experiments as a representative of typical continuously undulating scenarios. The height variation of this terrain follows a sinusoidal function. The robot moves through these terrain profiles at 0.5 m/s while keeping a constant desired leg length $L_{\text{leg},d}$. 

As shown in Figure \ref{fig:exp_data_3}, the vertical displacement and velocity of the robot's head along the z-axis, as well as the concurrent changes in contact forces during Experiment 3, are presented. The experimental results are summerized in \tabref{tab:stability_metrics_3}, the experimental results demonstrate that the proposed method achieves: (1) 49.2\% improved steady-state position accuracy 
, (2) 49.1\% better dynamic error suppression
, and (3) 38.8\% enhanced extreme disturbance adaptation
, while improving velocity tracking precision (velocity MAE reduction from 0.0416 m/s to 0.0287 m/s, 31.0\%) with 33.7\% smoother transient responses
and 14.2\% lower peak velocity variations
, collectively confirming superior terrain adaptation performance.

\begin{table}[h]
\centering
\caption{Stabilization performance metrics on experiment 3.}
\label{tab:stability_metrics_3}
\begin{tabular}{lccccc}
\toprule
\textbf{Metric} & \textbf{Unit} & \textbf{Baseline} & \textbf{Proposed} & \textbf{Improvement} \\
\midrule
\multicolumn{5}{l}{\textit{Position (z)}} \\
MAE & m & 0.0179 & \textbf{0.0091} & \textbf{49.2\%} \\
RMSE & m & 0.0224 & \textbf{0.0114} & \textbf{49.1\%} \\
P2P & m & 0.0849 & \textbf{0.0520} & \textbf{38.8\%} \\

\multicolumn{5}{l}{\textit{Velocity ($v_z$)}} \\
MAE & m/s & 0.0416 & \textbf{0.0287} & \textbf{31.0\%} \\
RMSE & m/s & 0.0725 & \textbf{0.0481} & \textbf{33.7\%} \\
P2P & m/s & 0.3861 & \textbf{0.3311} & \textbf{14.2\%} \\
\bottomrule
\end{tabular}
\end{table}

\section{CONCLUSION AND FUTURE WORK}\label{sec:conclusion}
In this paper, we propose an admittance control framework tailored to a 6-DoF wheeled bipedal robot, complemented by an online ground contact force estimation algorithm. Specifically, we propose a 2-DoF planar leg model for contact force estimation, while a W-SDIP model is put forward for the admittance control. Simulation experiments 
demonstrate that the proposed approach achieves enhanced 
steady-state position accuracy (MAE), achieves better dynamic error suppression (RMSE), and strengthens extreme disturbance adaptation (P2P),
compared to the baseline controller.
Future work will focus on deploying the algorithm to the physical DIABLO prototype for real-world validation, while extending the framework to a full 3D dynamic model to further boost terrain adaptability.

\bibliography{ROBIO} 

@string{icra = {Proc. of the {IEEE} Intl. Conf. on Robot. and Autom.}}

@string{iros = {Proc. of the {IEEE/RSJ} Intl. Conf. on Intell. Robots and Syst.}}

@misc{liu2024diablo6dofwheeledbipedal,
      title={DIABLO: A 6-DoF Wheeled Bipedal Robot Composed Entirely of Direct-Drive Joints}, 
      author={Dingchuan Liu and Fangfang Yang and Xuanhong Liao and Ximin Lyu},
      year={2024},
      eprint={2407.21500},
      archivePrefix={arXiv},
      primaryClass={cs.RO},
      url={https://arxiv.org/abs/2407.21500}, 
}

@online{BD2017handle,
author={Boston Dynamics},
title={Introducing Handle},
URL = {https://www.youtube.com/watch?v=-7xvqQeoA8c},
year={2019}
}

@INPROCEEDINGS{Jo2020,
  author={Jo, Joonhee and Oh, Yonghwan},
  booktitle={2020 IEEE/RSJ International Conference on Intelligent Robots and Systems (IROS)}, 
  title={Impedance Control of Humanoid Walking on Uneven Terrain With Centroidal Momentum Dynamics Using Quadratic Programming}, 
  year={2020},
  volume={},
  number={},
  pages={3525-3530},
  doi={10.1109/IROS45743.2020.9340799}}

@INPROCEEDINGS{Tsai2022,
  author={Tsai, Ching-Chih and Hsu, Wei-Ting and Tai, Feng-Chun and Chen, Shih-Che},
  booktitle={2022 International Automatic Control Conference (CACS)}, 
  title={Adaptive Motion Control of a Terrain-Adaptive Self-Balancing Leg-Wheeled Mobile Robot over Rough Terrain}, 
  year={2022},
  volume={},
  number={},
  pages={1-6},
  doi={10.1109/CACS55319.2022.9969857}}

@ARTICLE{Klemm2020,
  author={Klemm, Victor and Morra, Alessandro and Gulich, Lionel and Mannhart, Dominik and Rohr, David and Kamel, Mina and de Viragh, Yvain and Siegwart, Roland},
  journal={IEEE Robotics and Automation Letters}, 
  title={LQR-Assisted Whole-Body Control of a Wheeled Bipedal Robot With Kinematic Loops}, 
  year={2020},
  volume={5},
  number={2},
  pages={3745-3752},
  doi={10.1109/LRA.2020.2979625}}

@ARTICLE{Chen2021,
  author={Chen, Hua and Wang, Bingheng and Hong, Zejun and Shen, Cong and Wensing, Patrick M. and Zhang, Wei},
  journal={IEEE Robotics and Automation Letters}, 
  title={Underactuated Motion Planning and Control for Jumping With Wheeled-Bipedal Robots}, 
  year={2021},
  volume={6},
  number={2},
  pages={747-754},
  doi={10.1109/LRA.2020.3047787}}

@ARTICLE{Cui2021,
  author={Cui, Leilei and Wang, Shuai and Zhang, Jingfan and Zhang, Dongsheng and Lai, Jie and Zheng, Yu and Zhang, Zhengyou and Jiang, Zhong-Ping},
  journal={IEEE Robotics and Automation Letters}, 
  title={Learning-Based Balance Control of Wheel-Legged Robots}, 
  year={2021},
  volume={6},
  number={4},
  pages={7667-7674},
  doi={10.1109/LRA.2021.3100269}}

@INPROCEEDINGS{zhang2022,
  author={Zhang, Jingfan and Wang, Shuai and Wang, Haitao and Lai, Jie and Bing, Zhenshan and Jiang, Yu and Zheng, Yu and Zhang, Zhengyou},
  booktitle={2022 IEEE/RSJ International Conference on Intelligent Robots and Systems (IROS)}, 
  title={An Adaptive Approach to Whole-Body Balance Control of Wheel-Bipedal Robot Ollie}, 
  year={2022},
  volume={},
  number={},
  pages={12835-12842},
  doi={10.1109/IROS47612.2022.9981985}}

@ARTICLE{zhang2023,
  
AUTHOR={Zhang, Jingfan  and Li, Zhaoxiang  and Wang, Shuai  and Dai, Yuan  and Zhang, Ruirui  and Lai, Jie  and Zhang, Dongsheng  and Chen, Ke  and Hu, Jie  and Gao, Weinan  and Tang, Jianshi  and Zheng, Yu },
         
TITLE={Adaptive optimal output regulation for wheel-legged robot Ollie: A data-driven approach},
        
JOURNAL={Frontiers in Neurorobotics},
        
VOLUME={Volume 16 - 2022},

YEAR={2023},

URL={https://www.frontiersin.org/journals/neurorobotics/articles/10.3389/fnbot.2022.1102259},

DOI={10.3389/fnbot.2022.1102259},

ISSN={1662-5218}}

@INPROCEEDINGS{Hogan1984,
  author={Hogan, Neville},
  booktitle={1984 American Control Conference}, 
  title={Impedance Control: An Approach to Manipulation}, 
  year={1984},
  volume={},
  number={},
  pages={304-313},
  doi={10.23919/ACC.1984.4788393}}

@INPROCEEDINGS{Zhangchao2019,
  author={Zhang, Chao and Liu, Tangyou and Song, Shuang and Meng, Max Q.-H.},
  booktitle={2019 IEEE International Conference on Robotics and Biomimetics (ROBIO)}, 
  title={System Design and Balance Control of a Bipedal Leg-wheeled Robot}, 
  year={2019},
  volume={},
  number={},
  pages={1869-1874},
  doi={10.1109/ROBIO49542.2019.8961814}}

@INPROCEEDINGS{LiuTangyou2019,
  author={Liu, Tangyou and Zhang, Chao and Song, Shuang and Meng, Max Q.-H.},
  booktitle={2019 IEEE International Conference on Robotics and Biomimetics (ROBIO)}, 
  title={Dynamic Height Balance Control for Bipedal Wheeled Robot Based on ROS-Gazebo}, 
  year={2019},
  volume={},
  number={},
  pages={1875-1880},
  doi={10.1109/ROBIO49542.2019.8961739}}

@article{
Marco2022,
author = {Marco Tranzatto  and Takahiro Miki  and Mihir Dharmadhikari  and Lukas Bernreiter  and Mihir Kulkarni  and Frank Mascarich  and Olov Andersson  and Shehryar Khattak  and Marco Hutter  and Roland Siegwart  and Kostas Alexis },
title = {CERBERUS in the DARPA Subterranean Challenge},
journal = {Science Robotics},
volume = {7},
number = {66},
pages = {eabp9742},
year = {2022},
doi = {10.1126/scirobotics.abp9742},
URL = {https://www.science.org/doi/abs/10.1126/scirobotics.abp9742},
eprint = {https://www.science.org/doi/pdf/10.1126/scirobotics.abp9742}}

@article{Chen2024,
author = {Lu Chen;Lipeng Chen;Xiangchi Chen;Haojian Lu;Yu Zheng;Jun Wu;Yue Wang;Zhengyou Zhang;Rong Xiong;},
title = {Compliance while resisting: A shear-thickening fluid controller for physical human-robot interaction},
journal = {The International Journal of Robotics Research},
volume = {43},
number = {11},
pages = {1731-1769},
year = {2024},
doi = {10.1177/02783649241234364},
 URL = {
       http://dx.doi.org/10.1177/02783649241234364
},
 eprint = {
          http://dx.doi.org/10.1177/02783649241234364
},
}

@ARTICLE{Yu2023,
  author={Yu, Jianqiao and Zhu, Zhangzhen and Lu, Junyuan and Yin, Sicheng and Zhang, Yu},
  journal={IEEE Robotics and Automation Letters}, 
  title={Modeling and MPC-Based Pose Tracking for Wheeled Bipedal Robot}, 
  year={2023},
  volume={8},
  number={12},
  pages={7881-7888},
  doi={10.1109/LRA.2023.3322084}}

@INPROCEEDINGS{Landi2017,
  author={Landi, Chiara Talignani and Ferraguti, Federica and Sabattini, Lorenzo and Secchi, Cristian and Fantuzzi, Cesare},
  booktitle={2017 IEEE International Conference on Robotics and Automation (ICRA)}, 
  title={Admittance control parameter adaptation for physical human-robot interaction}, 
  year={2017},
  volume={},
  number={},
  pages={2911-2916},
  doi={10.1109/ICRA.2017.7989338}}

@inproceedings{klemm2019ascento,
  title={Ascento: A two-wheeled jumping robot},
  author={Klemm, Victor and Morra, Alessandro and Salzmann, Ciro and Tschopp, Florian and Bodie, Karen and Gulich, Lionel and K{\"u}ng, Nicola and Mannhart, Dominik and Pfister, Corentin and Vierneisel, Marcus and others},
  booktitle={2019 International Conference on Robotics and Automation (ICRA)},
  pages={7515--7521},
  year={2019},
  organization={IEEE}
}

@inproceedings{wang2021balance,
  title={Balance control of a novel wheel-legged robot: Design and experiments},
  author={Wang, Shuai and Cui, Leilei and Zhang, Jingfan and Lai, Jie and Zhang, Dongsheng and Chen, Ke and Zheng, Yu and Zhang, Zhengyou and Jiang, Zhong-Ping},
  booktitle={2021 IEEE International Conference on Robotics and Automation (ICRA)},
  pages={6782--6788},
  year={2021},
  organization={IEEE}
}
\end{document}